\pdfoutput=1

\documentclass[11pt]{article}

\usepackage[final]{acl}

\usepackage{times}
\usepackage{latexsym}

\usepackage[T1]{fontenc}

\usepackage[utf8]{inputenc}

\usepackage{microtype}
\usepackage{booktabs}
\usepackage{multirow}
\usepackage{geometry}
\usepackage{CJKutf8}
\usepackage{graphicx}
\usepackage{color}
\usepackage{multirow}
\usepackage{bm}
\usepackage{bbding}
\usepackage{array}
\usepackage{subfigure}
\usepackage{amsmath}
\usepackage{CJKutf8}
\usepackage{footmisc}
\title{PPT: Pre-trained Prompt Tuning for Few-shot Learning}

\author{
  Yuxian Gu$^{1,3*}$, 
  Xu Han$^{2,3*}$,
  Zhiyuan Liu$^{2,3,4}$,
  Minlie Huang$^{1,3,4\dagger}$\\
  $^1$The CoAI group, Tsinghua University, Beijing, China \\
  $^2$The THUNLP group, Tsinghua University, Beijing, China \\
  $^3$Institute for Artificial Intelligence, State Key Lab of Intelligent Technology and Systems, \\
Beijing National Research Center for Information Science and Technology, \\
Department of Computer Science and Technology, Tsinghua University, Beijing, China \\
  $^4$ Beijing Academy of Artificial Intelligence, BAAI, Beijing, China \\
  \texttt{\{guyx21,hanxu17\}@mails.tsinghua.edu.cn}\\
  \texttt{\{liuzy,aihuang\}@tsinghua.edu.cn}\\
 }

\begin{document}
\maketitle

\begin{CJK*}{UTF8}{gbsn}

\begin{abstract}

Prompts for pre-trained language models (PLMs) have shown remarkable performance by bridging the gap between pre-training tasks and various downstream tasks. Among these methods, prompt tuning, which freezes PLMs and only tunes soft prompts, provides an efficient and effective solution for adapting large-scale PLMs to downstream tasks. However, prompt tuning is yet to be fully explored. In our pilot experiments, we find that prompt tuning performs comparably with conventional full-model tuning when downstream data are sufficient, whereas it is much worse under few-shot learning settings, which may hinder the application of prompt tuning. We attribute this low performance to the manner of initializing soft prompts. Therefore, in this work, we propose to pre-train prompts by adding soft prompts into the pre-training stage to obtain a better initialization. We name this \textbf{P}re-trained \textbf{P}rompt \textbf{T}uning framework ``\textbf{\textsc{PPT}}''. To ensure the generalization of PPT, we formulate similar classification tasks into a unified task form and pre-train soft prompts for this unified task. Extensive experiments show that tuning pre-trained prompts for downstream tasks can reach or even outperform full-model fine-tuning under both full-data and few-shot settings. Our approach is effective and efficient for using large-scale PLMs in practice. The code is publicly available at \url{https://github.com/thu-coai/PPT}.

\end{abstract}

\section{Introduction}

{\let\thefootnote\relax\footnotetext{
$^\dagger$ Corresponding author. }
\let\thefootnote\relax\footnotetext{
$^*$ indicates equal contribution. }
}

\begin{figure*}[t]
    \centering
    \includegraphics[width=0.9\linewidth]{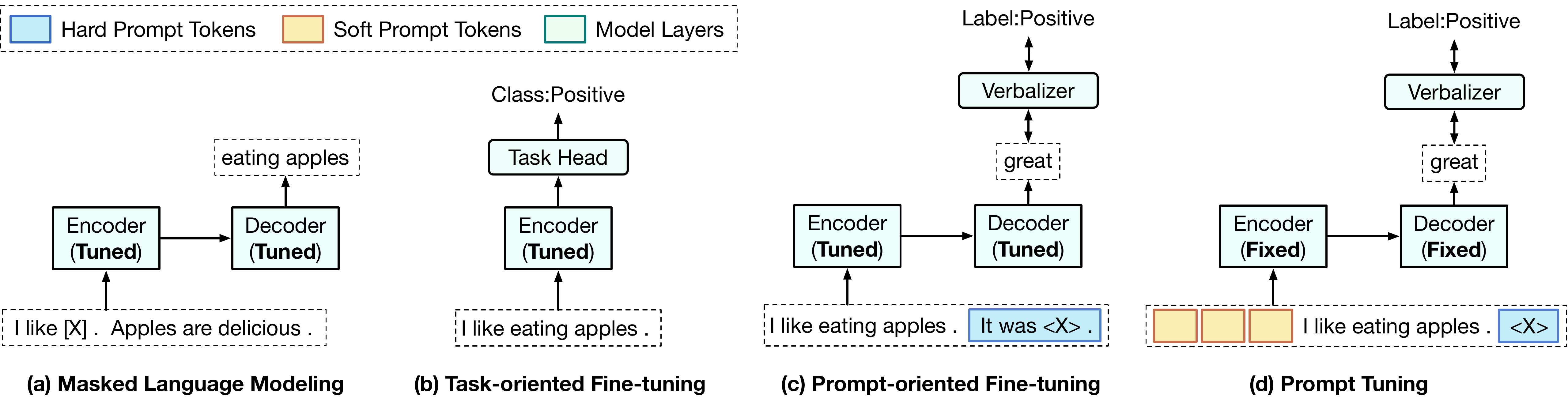}
    \caption{Paradigms of pre-training (masked language modeling), full-model tuning (task-oriented fine-tuning and prompt-oriented fine-tuning), and prompt tuning. The verbalizer is a function to map task labels to concrete words.〈X〉means the mask of typical pre-trained encoder-decoder models}
    \label{fig:example}
\end{figure*}

Fine-tuning pre-trained language models (PLMs)~\cite{bert,gpt2,t5} has made great progress in recent years. By tuning the entire model parameters, the versatile knowledge acquired from large-scale unlabeled corpora can be adapted to handling various NLP tasks and outperform the approach of learning models from scratch~\cite{plmsurvey}. For simplicity, we name this full-model tuning as ``\textsc{FT}''. As shown in Figure \ref{fig:example} (b) and (c), there are two mainstream \textsc{FT} approaches. The first one is task-oriented fine-tuning, where a task-specific head is added on top of PLMs, and the entire model is then fine-tuned by optimizing task-specific objectives on corresponding training data.

The second one is prompt-oriented fine-tuning~\cite{pet}, which is inspired by the recent works utilizing language prompts to probe the knowledge in PLMs~\cite{lama,gpt3}. In prompt-oriented fine-tuning, data samples are converted to sequences containing prompt tokens, and downstream tasks are formalized as language modeling problems. As shown in Figure~\ref{fig:example} (c), by adding the prompt ``It was $\left\langle \text{X} \right\rangle$ .'' to a sentence, we can determine its sentiment polarity with PLMs by predicting ``great'' or ``terrible'' at the mask position. As shown in Figure \ref{fig:example}, compared to task-oriented fine-tuning, prompt-oriented fine-tuning is more similar to the pre-training objectives (masked language modeling), thereby helping to better use knowledge in PLMs and often obtaining better performance. 

\begin{figure}[t]
    \subfigbottomskip=0cm
    \subfigure[Full-Data] { \label{fig:setting_a} 
    \centering
    \includegraphics[width=0.9\linewidth]{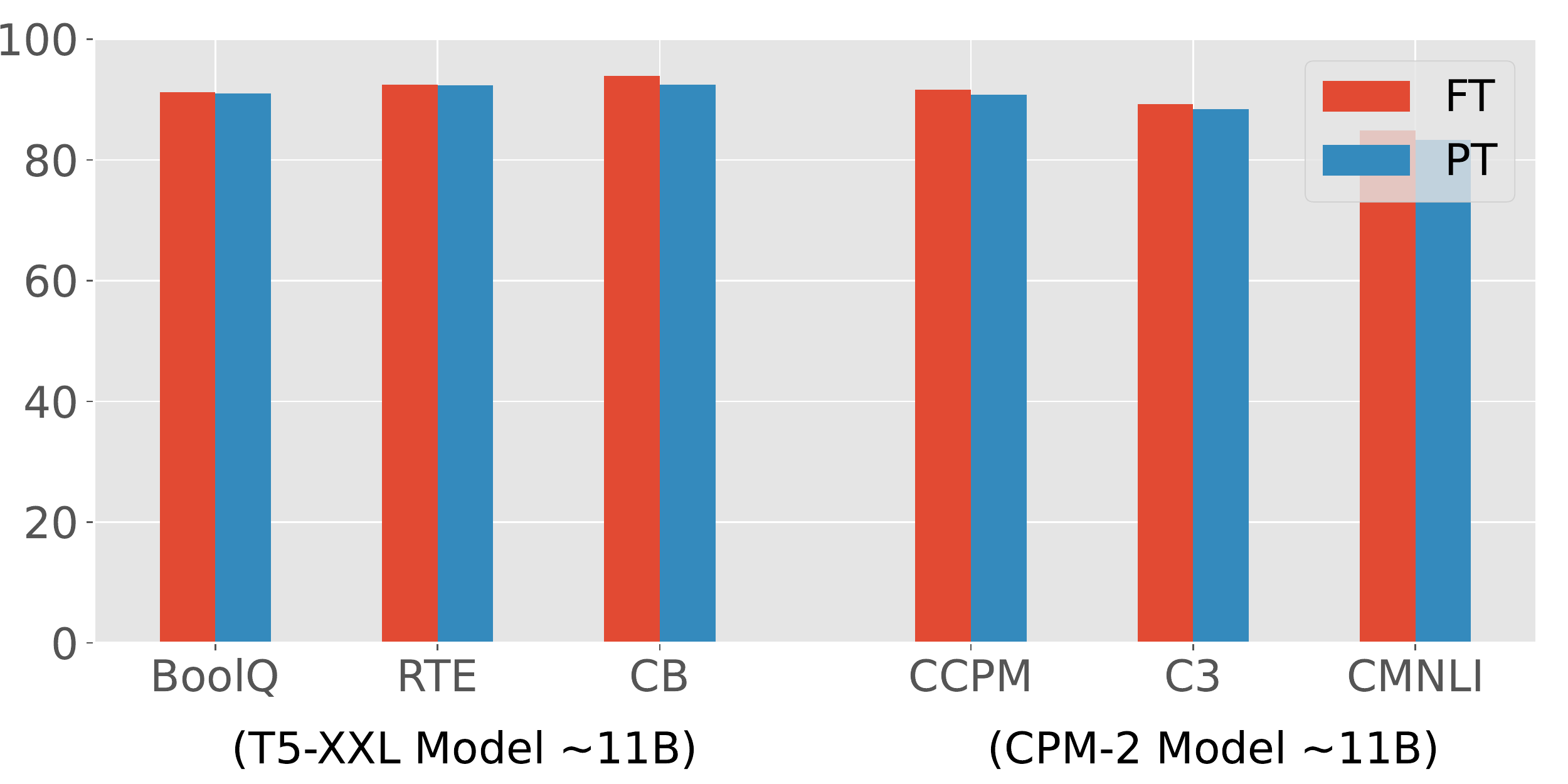} 
    }
    \subfigure[Few-Shot] { \label{fig:setting_b} 
    \centering
    \includegraphics[width=0.9\linewidth]{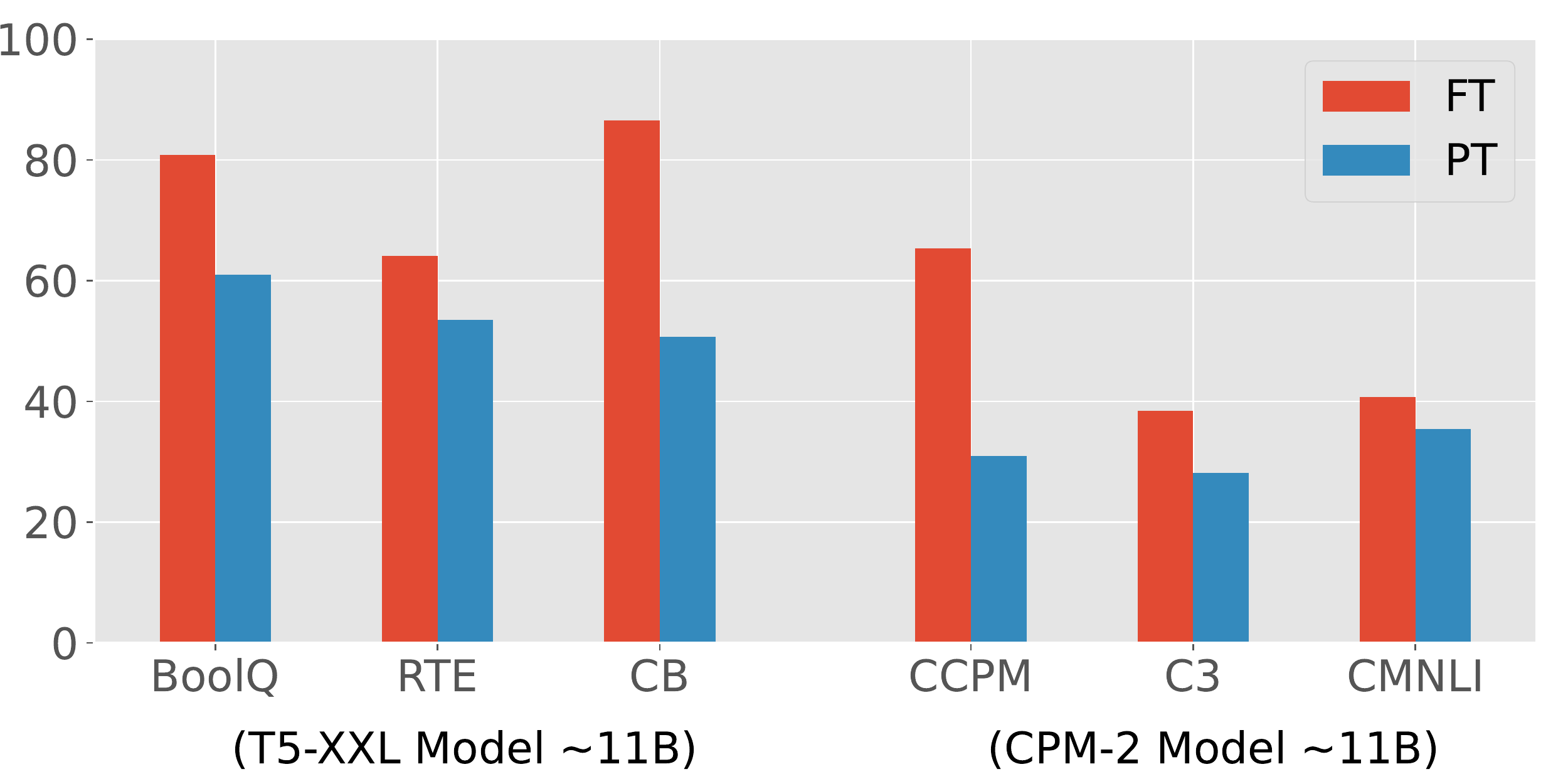} 
    } 
    \caption{Comparison between \textsc{PT} and \textsc{FT}. The tuned prompt is composed of 100 learnable embeddings whose dimensions are the same as the token embeddings of PLMs (4096 dimensions). All these results are based on 11B PLMs \textsc{T5} and \textsc{CPM-2}. \textsc{FT} needs to optimize all 11B parameters, while \textsc{PT} only trains about 410K prompt parameters.} 
    \vspace{-0.5em}
    \label{fig:setting} 
\end{figure} 

Although \textsc{FT} has shown promising results, with the rapid growth of model scale, fine-tuning and storing the entire large model for each downstream task becomes much more expensive. To address this challenge, \citet{prompt_tuning} proposes prompt tuning (\textsc{PT}) to adapt large PLMs to downstream tasks cheaply, as shown in Figure~\ref{fig:example} (d). Specifically, \textsc{PT} uses soft prompts composed of continuous embeddings instead of hard prompts (discrete language phrases). These continuous prompts are generally randomly initialized and learned end-to-end. To avoid storing the entire model for each downstream task, \textsc{PT} freezes all PLM parameters and merely tunes soft prompts, without adding any intermediate layers and task-specific components. 

\textsc{PT} has two promising advantages. First, soft prompts can be learned end-to-end in comparison to hard prompts. Second, \textsc{PT} is an efficient and effective paradigm for the practical use of large-scale PLMs, which is comparable to \textsc{FT} when downstream data are sufficient (Figure \ref{fig:setting_a}). However, as shown in Figure \ref{fig:setting_b}, we find that \textsc{PT} performs much worse than \textsc{FT} under few-shot settings, which may hinder the application of \textsc{PT} in various low-resource scenarios.

Hence, in this paper, we explore how to use PLMs for few-shot learning in an efficient and effective manner through \textsc{PT}. Specifically, we conduct pilot experiments to empirically analyze the effectiveness of \textsc{PT} on PLMs in Section \ref{sec:pilot}, which is ignored by most existing works. Our discoveries are as follows: (1) the verbalizer choice has a large impact on the performance; (2) simply initializing soft prompts with concrete word embeddings fails to improve the performance, yet (3) combining soft and hard prompts is helpful; and (4) all these methods cannot handle few-shot prompt tuning problems well. The above observations reveal that prompt searching for PLMs is not trivial, and carefully initialized soft prompt tokens is crucial. 

To help the model find suitable prompts, we pre-train these tokens with self-supervised tasks on large-scale unlabeled corpora. To ensure the generalization of pre-trained prompts, we group typical classification tasks into three formats: sentence-pair classification, multiple-choice classification, and single-text classification, each format corresponding to one self-supervised pre-training task. In addition, we find multiple-choice classification more general among these formats and we can unify all classification tasks to this format. We name this \textbf{P}re-trained \textbf{P}rompt \textbf{T}uning framework ``\textbf{\textsc{PPT}}''. We evaluate \textsc{PPT} on several datasets based on three 11B PLMs: T5-XXL~\cite{t5}, mT5-XXL~\cite{mt5} and CPM-2~\cite{cpm-v2} in few-shot scenarios. Experiments show that \textsc{PPT} can not only improve \textsc{PT} by a large margin, reaching or even outperforming \textsc{FT} methods, but also reduce the variance of few-shot learning. Besides the effectiveness, \textsc{PPT} also retains the parameter efficiency of \textsc{PT}, which is valuable for future applications on large-scale PLMs.

\begin{table}[t]
    \footnotesize
    \centering
    \begin{tabular}{@{}p{14em}p{4em}l@{}}
    \toprule
    Hard Prompt                                                                  & Verbalizer     & Accuracy  \\ \midrule
    None                                                    & good/bad      & $70.5_{15.5}$ \\
    Man \#1: $\bm{P}$ $s$. It was $\left\langle \text{X}\right\rangle$.     & good/bad       & $87.6_{6.6}$ \\
    Man \#2: $\bm{P}$ Just $\left\langle \text{X}\right\rangle$ ! $s$    & good/bad       & $86.0_{8.1}$   \\
    Man \#3: $\bm{P}$ $s$. All in all, it was $\left\langle \text{X}\right\rangle$.                   & good/bad       & $83.4_{8.3}$  \\ \midrule
    Gen \#1: $\bm{P}$ .$s$. a $\left\langle \text{X}\right\rangle$.   & good/bad       & $81.6_{13.8}$   \\
    Gen \#2: $\bm{P}$ $s$. \ A $\left\langle \text{X}\right\rangle$ one. & good/bad       & $81.2_{2.2}$  \\ \midrule
    Man \#1: $\bm{P}$ $s$. \ It was $\left\langle \text{X}\right\rangle$.     & great/terrible & $86.9_{7.9}$    \\
    Man \#1: $\bm{P}$ $s$. \ It was $\left\langle \text{X}\right\rangle$.     & dog/cat        & $60.0_{7.6}$     \\
    Man \#1: $\bm{P}$ $s$. \ It was $\left\langle \text{X}\right\rangle$.     & bad/good       & $76.3_{11.7}$    \\ \midrule
    Full-Model Tuning                                                           & good/bad         & \bm{$91.4_{0.8}$}    \\
    \bottomrule
    \end{tabular}
    \caption{The impact of hard prompts and verbalizers on \textsc{PT} for few-shot learning (32 samples) on SST-2.  $\bm{P}$ represents soft prompts. $s$ denotes the input sentence. ``Man'' means manually designed hard prompts and ``Gen'' means auto-generated hard prompts. The choice of hard prompts and verbalizers has a significant influence on model performance. }
    \label{tab:hybrid}
\end{table}

\section{Pilot Experiments}
\label{sec:pilot}

In this section, we present pilot experiments of \textsc{PT} for few-shot learning. We analyze three strategies including hybrid prompt tuning, verbalizer selection, and real word initialization. We follow \citet{prompt_tuning} to test \textsc{PT} with T5-XXL (11B parameters) and use 100 tunable soft prompt tokens\footnote{Using 100 soft prompt tokens achieves the best performance in \citet{prompt_tuning}.}. 

Following \citet{pet2}, we randomly select 32 samples to construct the training set $D_{\text{train}}$ from the original training data. To tune the hyper-parameters, we compose a validation set $D_{\text{dev}}$ from the original training data and ensure $|D_{\text{train}}| = |D_{\text{dev}}|$ to simulate the few-shot learning setting~\cite{true-few-shot}. We follow \citet{few-sample-bert} and \citet{lm-bff} to use the original validation set as the test set $D_{\text{test}}$, which means $|D_{\text{test}}| \gg |D_{\text{train}}| = |D_{\text{dev}}|$.

\paragraph{Hybrid Prompt Tuning}
\label{sec:hybrid_pt}
In hybrid prompt tuning, both soft and hard prompts are used~\cite{p-tuning,ptr}. However, previous works train soft prompts jointly with the entire model. In \textsc{PT} where only prompt tokens are tunable, the effectiveness of hybrid prompts is under-explored. In Table \ref{tab:hybrid}, we show the results of combining soft prompts $\bm{P}$ with three manually designed hard prompts and two auto-generated hard prompts~\cite{lm-bff} on a sentiment classification task~\cite{sst-2}. We can see that hard prompts improve \textsc{PT}, but still under-perform \textsc{FT}. Furthermore, different hard prompts affect the performance remarkably, therefore much human labor for prompt design and selection is needed. 

\paragraph{Verbalizer Selection}
Verbalizer maps task-specific labels to concrete tokens. For instance, in Figure \ref{fig:example} (c) and (d), the verbalizer maps  the label ``Positive'' to ``great''. From Table \ref{tab:hybrid} we can see that the choices of verbalizers influence the performance remarkably. In general, common words that explain the meaning of corresponding labels work well. This also guides our verbalizer selection for \textsc{PPT} in Section \ref{sec:ppt}.

\paragraph{Real Word Initialization}
In real word initialization, we use the embeddings of concrete words to initialize the soft prompt and test four initialization strategies. The effectiveness of this approach has been verified on small PLMs (fewer than 3B parameters) in previous works~\cite{prompt_tuning}. However, from the experiments on SST-2~\cite{sst-2} and BoolQ~\cite{boolq} (Table \ref{tab:init}), we find that for the 11B model, real word initialization has little or even negative impact on the performance in few-shot scenarios. This suggests that observations on small models can not be directly adapted to large models and finding a good initialization for soft prompts is yet to be explored.

\begin{table}[t]
    \centering
    \small
    \begin{tabular}{lll}
    \toprule
                     & SST-2 & BoolQ\\
    \midrule
        Random Init. & $70.5_{15.5}$ & $61.0_{5.3}$ \\
        Label Init. & $58.9_{2.7}$ & $63.0_{0.4}$ \\
        Vocab Sampling & $57.0_{4.0}$ & $58.4_{4.9}$ \\
        Top-1000 Sampling & $57.9_{4.2}$ & $57.7_{3.9}$ \\
        Task-Related Sampling & $58.5_{3.8}$ & $58.2_{4.0}$ \\
    \midrule
        Full-Model Tuning & \bm{$91.4_{0.8}$} & \bm{$80.8_{2.4}$} \\ 
    \bottomrule
    \end{tabular}
    \caption{Few-shot learning performance with different strategies for choosing concrete words for prompt initialization in \textsc{PT}. ``Label Init'': use the embeddings of the label words. ``Vocab Sampling'': randomly sample words from the vocabulary. ``Top-1000 Sampling'': randomly sample words from the most frequent 1000 words in the pre-training corpus. ``Task-Related'': randomly sample words from the downstream data. We use the classification accuracy (\%) for evaluation.}
    \label{tab:init}
\end{table}

To summarize, although the above enhancement strategies cannot help \textsc{PT} achieve comparable results with \textsc{FT} under few-shot settings, they are still the key factors that influence the \textsc{PT} performance. In the following sections, we describe our \textsc{PPT} framework and show in experiments that \textsc{PPT} not only provides a good prompt initialization, but also takes advantage of the good verbalizer, and is complementary to hybrid prompts.

\section{Pre-trained Prompt Tuning (PPT)}
\label{sec:ppt}

In this section, we describe the whole framework of \textsc{PPT}, including how to pre-train prompts and use these pre-trained prompts for specific tasks. 

\subsection{Overview}

Following the approach of T5~\cite{t5} and \textsc{PT}~\cite{prompt_tuning}, we solve all downstream tasks in a text-to-text format. As shown in Figure \ref{fig:example} (c), to reduce the objective gap between pre-training and downstream tasks, prompt-oriented fine-tuning converts downstream tasks into cloze-style objectives. Taking classification for example, given an input sentence $\bm{x} \in \mathcal{V}^*$ and its label $y \in \mathcal{Y}$, a pattern mapping $f: \mathcal{V}^* \mapsto \mathcal{V}^*$ is first applied to convert $\bm{x}$ into a new sequence $f(\bm{x})$, where $\mathcal{V}$ is the vocabulary of PLMs. $f(\bm{x})$ not only adds some prompt tokens as hints, but also preserves the mask token $\left\langle \text{X} \right\rangle$ to let PLMs predict tokens at the masked positions. Then, a verbalizer $v: \mathcal{Y} \mapsto \mathcal{V}^*$ is used to map $y$ to some label tokens $v(y)$. With $f(\cdot)$ and $v(\cdot)$, a classification task can be represented by a pattern-verbalizer pair $(f, v)$:
\begin{equation}
\small
\label{eq:prompt_fine_tuning}
\begin{aligned}
&\arg\max_{\bm{\theta}} \sum_{\bm{x}} \log p\big(y|\bm{x};\bm{\theta}\big) \\
&= \arg\max_{\bm{\theta}} \sum_{\bm{x}} \log p\big(\left\langle \text{X} \right\rangle = v(y) | f(\bm{x});\bm{\theta}\big),
\end{aligned}
\end{equation}
where $\bm{\theta}$ indicates all tunable parameters, especially the parameters of PLMs. For convenience, we use ``$\text{PVP}$'' to denote this pattern-verbalizer pair~\cite{pet}. 

In \textsc{PT}~\cite{prompt_tuning}, a set of soft prompts $\bm{P}$ are concatenated to the beginning of the sequence and the model input becomes $[\bm{P}; f(\bm{x})]$,
where $[\cdot;\cdot]$ is the concatenation operation. By tuning $\bm{P}$ , Eq.~(\ref{eq:prompt_fine_tuning}) is replaced by
\begin{equation}
\small
\label{eq:prompt_tuning}
\arg\max_{\bm{P}} \sum_{\bm{x}} \log p\big( \left\langle \text{X} \right\rangle =v(y) \mid [\bm{P};f(\bm{x})] ; \bm{P}\big).
\end{equation}
Owing to the power of large-scale PLMs, Eq.~(\ref{eq:prompt_tuning}) is verified to be comparable to these \textsc{FT} methods under full-data settings. However, we find it hard to learn effective soft prompts, which may result in low performance in various few-shot scenarios. The parameter initialization usually has a large impact on the difficulty of the model training and optimization, and our pilot experiments have shown that existing initialization strategies have little or even negative impact on the \textsc{PT} performance of large-scale PLMs. We refer more details of these pilot experiments to Section~\ref{sec:exp}. 

\begin{figure}
    \includegraphics[width=\linewidth]{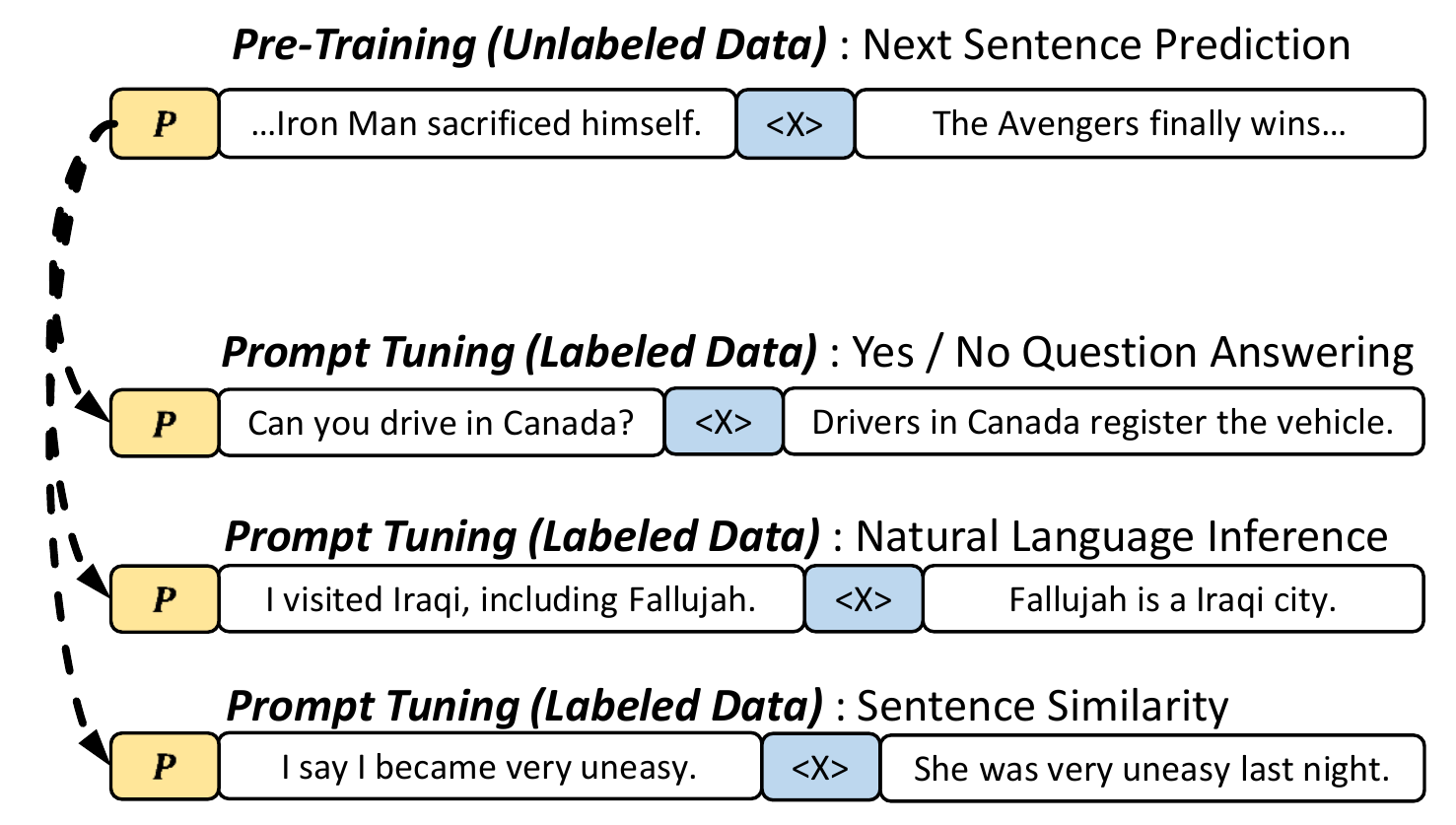}
    \caption{An example of \textsc{PPT} used in sentence pair tasks. $P$ denotes soft prompt. $\left\langle \text{X}\right\rangle$ means the mask of typical encoder-decoder model like T5 and CPM-2.}\label{fig:nsp}
\end{figure}

Recently, pre-training has been proven to be an effective method to find a good model initialization. Inspired by this, we propose to pre-train soft prompts. We notice that some groups of downstream tasks are related to certain self-supervised tasks built on unlabeled pre-training corpora. For instance, some tasks in the form of sentence-pair classification, such as natural language inference and sentence similarity, are similar to the next sentence prediction (NSP)~\cite{bert} task used in the pre-training stage. As shown in Figure \ref{fig:nsp}, these tasks all take two sentences as input and compare their semantic meanings. Therefore, soft prompts pre-trained by NSP can be a good initialization for these sentence-pair tasks. 

Formally, suppose we can divide downstream tasks into $m$ groups $\{ \mathcal{T}_1, \mathcal{T}_2, ..., \mathcal{T}_m \}$, where $\mathcal{T}_i$ is the set containing $n_i$ downstream tasks: $\{\text{PVP}^{1}_i, \text{PVP}^{2}_i, ..., \text{PVP}^{n_i}_{i}\}$, where $\text{PVP}^{k}_i = (f^k_{i}, v^k_{i})$. For each group, we design a corresponding pre-training task $\text{PVP}^{\text{pre}}_i = (f^{\text{pre}}_{i}, v^{\text{pre}}_{i})$. After pre-training soft prompts on these tasks with all model parameters fixed, we get $m$ pre-trained prompts $ \{ \bm{P}_1, \bm{P}_2, ..., \bm{P}_m\}$. Then, for each task $\text{PVP}^{k}_i$ in $\mathcal{T}_i$, we continue to optimize Eq.~(\ref{eq:prompt_tuning}) by using $\bm{P}_i$ as the soft prompts initialization.

\subsection{Designing Pattern-Verbalizer Pairs for Pre-training}

In this section, we take three typical classification tasks as examples to describe the design of pattern-verbalizer pairs $\text{PVP}^{\text{pre}}_i$ for prompt pre-training. 

\subsubsection{Sentence-Pair Classification}

Sentence-pair classification tasks such as natural language inference and sentence similarity take two sentences $\bm{x}=(\bm{s}_1, \bm{s}_2)$ as the input. To design a $\text{PVP}$ for these tasks, we extend the next sentence prediction in \citet{bert} to a 3-class classification with labels $\mathcal{Y} = \{ 0, 1, 2 \}$ as the pre-training task. These labels in $\mathcal{Y}$ can respectively indicate that the semantic relation between two sentences is coherent (with label 2), similar (1) and irrelevant (0). To construct signal from unlabeled documents, we set the two sentences next to each other as label 2, those from the same document but not true next sentences as 1, and those from different documents as 0. We consider the label set $|\mathcal{Y}| \leq 3$ because this covers most sentence pair tasks. $\text{PVP}^{\text{pre}}_i = (f^{\text{pre}}_i, v^{\text{pre}}_i)$ is given as
\begin{equation}
    \small
    \begin{aligned}
        f^{\text{pre}}_i(\bm{x}) &= \text{``}\bm{s}_1 \left\langle \text{X} \right\rangle . \bm{s}_2\text{''}, \\
        v^{\text{pre}}_i(\mathcal{Y}) &= [\text{no}, \text{maybe}, \text{yes}].
    \end{aligned}
\end{equation}
Designing $\text{PVP}^{k}_{i}=( f^k_{i}, v^k_{i} )$ according to $\text{PVP}^{\text{pre}}_i$ is simple. $\bm{s}_1$ and $\bm{s}_2$ can be replaced by the input sentence pair. If a task outputs two labels, then we take $v^k_i(\mathcal{Y}) = [\text{no}, \text{yes}]$. If a task outputs three labels, we set $v^k_{i} = v^{\text{pre}}_i$. If a task requires to measure the similarity between two sentences, the probability over $\{\text{no}, \text{yes}\}$ can serve for this task.

\subsubsection{Multiple-Choice Classification}
\label{sec:mc}
Many tasks can be formulated as multiple-choice classification, which takes a query and several answer candidates as the input. We design a next sentence selection task to pre-train the prompt. Given a sentence as the query $\bm{s}_q$, the model is trained to select the adjacent sentence from six candidates, denoted as $\bm{s}_1 \sim \bm{s}_6$ and thus the label set is $\mathcal{Y}=\{ 1,2,3,4,5,6 \}$. These candidates consist of the right answer, one sentence from the same document but is not adjacent to the query, and four sentences from other documents. For $\bm{x} = (\bm{s}_q, \bm{s}_1, \bm{s}_2, \cdots, \bm{s}_6)$, $( f^{\text{pre}}_i, v^{\text{pre}}_i )$ is given as 
\begin{equation}
    \small 
    \begin{aligned}
        f^{\text{pre}}_i(\bm{x}) &= \text{``}\bm{s}_q\text{? A.}\bm{s}_1 \cdots \text{F.}\bm{s}_6. \text{Answer is} \left\langle \text{X}\right\rangle.\text{''}, \\
        v^{\text{pre}}_i(\mathcal{Y}) &= [\text{A}, \text{B}, \text{C}, \text{D}, \text{E}, \text{F}].
    \end{aligned}
\end{equation}
Most multiple-choice tasks can use $\{ f^{\text{pre}}_i, v^{\text{pre}}_i \}$ directly as their $\text{PVP}$s. For tasks like reading comprehension, the input may contain a passage and a question. We concatenate them to form the query.

\subsubsection{Single-Sentence Classification}
\label{sec:ssc}

For single-sentence classification, we create pseudo labels for prompt pre-training. Taking sentiment classification as an example, we use another small model to annotate sentiment labels for the sentences from the pre-training corpus and filter out those with low classification probability. In practice, we use a RoBERTa$_\text{BASE}$~\cite{roberta} model fine-tuned on a 5-class sentiment classification dataset other than the few-shot datasets we evaluate on. Then with a sentence $\bm{s}$ from the corpus, we have the input $\bm{x}=(\bm{s})$ and the label set $\mathcal{Y}=\{1,2,3,4,5\}$. $( f^{\text{pre}}_i, v^{\text{pre}}_i )$ is given as
\begin{equation}
    \small
    \begin{aligned}
        f^{\text{pre}}_i(\bm{x}) &= \text{``}\bm{s}.\left\langle \text{X}\right\rangle.\text{''},  \\
        v^{\text{pre}}_i(\mathcal{Y}) &= [\text{terrible}, \text{bad}, \text{maybe}, \text{good}, \text{great}].
    \end{aligned}
\end{equation}
For sentiment classification tasks with 5 labels, we can use $\text{PVP}^{k}_{i} = \text{PVP}^{\text{pre}}_i$. For those  with fewer than 5 labels, we choose a subset from $v^{\text{pre}}_i(\mathcal{Y})$ as labels. 

Although the above method improves the model performance, we have to point out that it is still limited to generalize to other single-text classifications in different domains and with different numbers of labels. Therefore, the method described in the following section is proposed to solve this problem.

\subsection{Unifying Task Formats}

The above-mentioned $\text{PVP}$s for pre-training can be unified to a single format: multiple-choice classification. Specifically, for sentence-pair classification, the query is the concatenation of the two sentences and there are three options: no, maybe, and yes. For single-sentence classification, the query is the input sentence and the options are the concrete labels. Note that in this way, the pre-trained $\text{PVP}$s can be used in single text classification tasks from arbitrary domains and with much more labels.

Constructing a unified $\text{PVP}$ is similar to the idea of MultiQA~\cite{multi-qa} and UnifiedQA~\cite{unified-qa}. Recently, \citet{meta-zero-shot} use some hard prompts to unify several tasks as a meta question answering task. They tune the entire model with this meta task on a collection of QA datasets and then transfer to other classification tasks under low-resource settings. However, our \textsc{PPT} focuses on tuning soft prompts with the main body of PLMs fixed and our pre-training is conducted on fully unsupervised data, rather than the collection of supervised datasets.

Since different tasks may have different candidate numbers and lengths, we construct pre-training samples with option numbers varying from 2 to 16~\footnote{We set $16$ labels in this paper as they can cover most benchmarks, but more labels are applicable for other tasks.} and option lengths from 50 to 20. We use the $\text{PVP}$ in Section \ref{sec:mc} for pre-training, and then apply pre-trained soft prompts to cover the above mentioned three classification tasks.

\begin{table}[t]
    \centering
    \footnotesize
    \begin{tabular}{@{}lll|lll@{}}
        \toprule
        \multicolumn{3}{c|}{English}  & \multicolumn{3}{c}{Chinese}              \\ \midrule
        Dataset     & Format & $n_{\text{class}}$ & Dataset                & Format & $n_{\text{class}}$ \\
        \midrule
        SST-2       & SSC       &   2      & ChnSent                &  SC     &    2     \\
        SST-5       & SSC       &   5      & Amazon                 &  SC     &    5     \\
        YahooAns    & SSC       &   10     & TNews                  &  SC     &    14     \\
        RACE-m      & MCC       &   4      & CCPM                   &  MCC     &    4     \\
        RACE-h      & MCC       &   4      & C$^3$ &  MCC     &    4     \\
        BoolQ       & SPC      &   3      & LCQMC                  &  SPC    &    3     \\
        RTE         & SPC      &   3      & CMNLI                  &  SPC    &    3     \\
        CB          & SPC      &   3      & OCNLI                  &  SPC    &    3     \\ \bottomrule
    \end{tabular}
    \caption{The datasets we evaluate. The ``Format'' column means the task category. SSC stands for single-sentence classification, MCC for multiple-choice classification, and SPC for sentence-pair classification. $n_{\text{class}}$ means the label number of each dataset.}
    \label{tab:dataset}
    \vspace{-1em}
\end{table}

\begin{table*}[t]
    \centering
    \footnotesize
    \begin{tabular}{p{0.897cm}<{\centering}lc|cc|cc|ccc}
    \toprule
    \multicolumn{10}{c}{English Tasks} \\ \midrule
                &\multirow{2}*{Model} & \multirow{2}*{Method} &   SST-2      & SST-5        &  RACE-m      &  RACE-h      & BoolQ        & RTE          & CB\\
                &                     &                       &     Acc.     &  Acc.        &  Acc.        &  Acc.        & Acc.         & Acc.         &  F1\\
    \midrule
    \multirow{4}*{\shortstack{FT\\(11B)}} 
                & T5-Small             & -                     & $72.8_{3.1}$ & $31.1_{0.4}$ & $26.4_{0.6}$ & $26.3_{0.5}$ & $59.2_{0.6}$ & $54.0_{1.7}$ & $70.1_{4.6}$ \\
                & T5-Base             & -                     & $74.6_{2.7}$ & $28.8_{1.8}$ & $27.2_{0.5}$ & $26.7_{0.2}$ & $61.9_{2.1}$ & $56.1_{2.3}$ & $70.4_{2.6}$ \\
                & T5-Large            & -                     & $89.1_{2.2}$ & $42.4_{1.2}$ & $48.2_{1.6}$ & $43.2_{1.7}$ & $74.6_{0.9}$ & $64.4_{3.4}$ & $82.3_{2.2}$ \\
                & T5-XL               & -                     & $89.6_{3.2}$ & $38.4_{5.1}$ & $55.0_{2.8}$ & $50.9_{2.6}$ & $77.2_{2.1}$ & $62.3_{6.8}$ & $81.9_{9.0}$ \\
                & T5-XXL              & -                     & $91.4_{0.8}$ & $40.6_{2.0}$ & \bm{$62.9_{3.9}$} & \bm{$54.8_{3.0}$} & $80.8_{2.4}$ & $64.1_{2.0}$ & \bm{$86.5_{5.3}$} \\
    \midrule
    \multirow{6}*{\shortstack{PT\\(410K)}} 
    &\multirow{6}*{T5-XXL}
                                      &   Vanilla PT           & $70.5_{15.5}$ & $32.3_{8.3}$ & $34.7_{8.2}$ & $31.6_{3.5}$ & $61.0_{5.3}$ & $53.5_{3.5}$ & $50.7_{4.1}$ \\
    &                                 &   Hybrid PT           & $87.6_{6.6}$ & $40.9_{2.7}$ &  $53.5_{8.2}$ & $44.2_{6.4}$ & $79.8_{1.5}$ & $56.8_{2.6}$ & $66.5_{7.2}$ \\
    &                    & LM Adaption                        & $77.6_{7.5}$ & $36.2_{3.6}$ & $27.3_{0.2} $ & $26.5_{0.4}$ & $62.0_{0.3}$ & $55.3_{1.0}$ &  $61.2_{1.7}$ \\
    \cmidrule(l){3-10}  
    &                                & PPT                    & $93.5_{0.3}$ & \underline{\bm{$50.2_{0.7}$}} & $60.0_{1.2}$  & \underline{$53.0_{0.4}$} & $66.43_{5.7}$ & $58.9_{1.6}$  & $71.2_{6.2}$ \\
    &                                & Hybrid PPT             & $93.8_{0.1}$ & $50.1_{0.5}$ & \underline{$62.5_{0.9}$} & $52.2_{0.7}$ & \underline{\bm{$82.0_{1.0}$}} & $59.8_{3.2}$ &  $73.2_{7.0}$ \\
    &                                & Unified PPT            & \underline{\bm{$94.4_{0.3}$}} & $46.0_{1.3}$ &  $58.0_{0.9}$ & $49.9_{1.3}$ & $76.0_{2.7}$ & \underline{\bm{$65.8_{2.1}$}} & \underline{$82.2_{5.4}$} \\
    
    \midrule[0.5pt]

    \multicolumn{10}{c}{Chinese Tasks} \\ \midrule
                & \multirow{2}*{Model} & \multirow{2}*{Method}  & ChnSent      & Amazon       & CCPM            & C$^3$           & LCQMC            & CMNLI            & OCNLI \\
                &                      &                        & Acc.         & Acc.         & Acc.            & Acc.            & Acc.             & Acc.             & Acc. \\
    \midrule
    \multirow{6}*{\shortstack{FT\\(11B)}}
                & mT5-Small        & -                          & $76.1_{2.6}$ & $29.9_{1.9}$ & $31.9_{1.2}$    & $29.6_{0.5}$    &   $52.4_{2.5}$   &  $36.5_{0.2}$    & $34.9_{1.3}$ \\
                & mT5-Base         & -                          & $78.2_{0.6}$ & $36.4_{0.9}$ & $40.4_{6.8}$    & $29.4_{0.6}$    &   $50.9_{1.0}$   &  $36.3_{0.5}$    & $35.4_{0.6}$ \\
                & mT5-Large        & -                          & $79.1_{0.6}$ & $31.0_{1.4}$ & $46.0_{4.0}$    & $29.9_{0.8}$    &   $52.1_{0.6}$   &  $35.8_{1.2}$    & $35.2_{1.1}$ \\
                & mT5-XL           & -                          & $82.7_{2.6}$ & $35.5_{1.7}$ & $68.3_{5.1}$    & $29.7_{1.2}$    &   $52.9_{2.4}$   &  $36.8_{1.6}$    & $35.6_{0.5}$ \\
                & mT5-XXL          & -                          & $83.6_{1.5}$ & $42.1_{0.8}$ & $79.7_{1.1}$    & $37.2_{3.3}$    &   $53.1_{1.0}$   &  $39.0_{0.4}$    & $37.4_{1.2}$ \\
                & CPM-2            & -                          & $86.1_{1.8}$ & $42.5_{2.0}$ & $81.8_{1.6}$    & $38.4_{3.7}$    &   $58.8_{1.8}$   &  $40.7_{1.0}$    & $38.5_{1.5}$ \\
    \midrule
    \multirow{6}*{\shortstack{PT\\(410K)}}&
    \multirow{6}*{CPM-2}
                                   &  Vanilla PT       & $62.1_{3.1}$ & $30.3_{4.8}$ & $31.0_{9.7}$  & $28.2_{0.4}$   & $51.5_{3.4}$      & $35.4_{0.5}$     & $37.0_{0.5}$ \\
    &                              &  Hybrid PT        & $79.2_{4.0}$  & $39.1_{3.8}$ & $46.6_{15.0}$  & $29.2_{0.5}$   & $54.6_{2.3}$      & $37.1_{0.6}$   & $37.8_{1.4}$ \\
    &                              & LM Adaption       & $74.3_{5.2}$ & $35.2_{2.4}$ & $33.7_{12.8}$ &  $30.2_{1.5}$   &  $51.4_{2.9}$    & $35.1_{0.3}$    & $38.0_{1.1}$ \\ \cmidrule(l){3-10}  
    &                              & PPT               & $90.1_{0.8}$ & $48.6_{0.6}$ & \underline{\bm{$85.4_{0.6}$}}  & $43.8_{2.2}$  &    $59.1_{0.6}$  & \underline{\bm{$43.0_{0.5}$}} &  $40.1_{0.4}$ \\
    &                              & Hybrid PPT        & $89.5_{0.3}$ & \underline{\bm{$48.8_{2.0}$}} & $83.9_{0.5}$ & $46.0_{0.5}$ &    \underline{\bm{$67.3_{0.9}$}}  & $41.3_{0.8}$ &  $38.7_{0.6}$ \\
    &                              & Unified PPT       & \underline{\bm{$90.7_{0.2}$}} & $44.6_{1.1}$ & $83.4_{0.9}$ & \underline{\bm{$50.2_{0.6}$}} & $55.0_{0.4}$ & $40.6_{0.4}$ & \underline{\bm{$41.5_{1.5}$}} \\
    \bottomrule
    \end{tabular}
    \caption{Classification results. The experiments are conducted with 32 training samples and 32 validation samples on each dataset. FT means full-model tuning, where the entire model (with about 11B parameters) should be tuned on each dataset. PT means prompt tuning, where only 410K parameters are trained. We report the mean and the standard deviation over 5 random seeds. The score marked as \textbf{bold} means the best performance among all the methods. The score marked with an \underline{underline} means the best one among prompt tuning (PT) methods.} \label{tab:exp_main}
    \vspace{-1em}
\end{table*}

\section{Experiments}


\label{sec:exp}
\subsection{Setup}
We conduct experiments on both Chinese and English tasks (see Table \ref{tab:dataset}). As described in Section \ref{sec:pilot}, for tasks with fewer than 5 labels, we construct $D_{\text{train}}$ and $D_{\text{dev}}$ with 32 samples from the original training data and ensure the number of labels is balanced. For tasks with more than 5 labels like TNews and YahooAnswer, it is hard to compose a dataset with label-balanced samples. Therefore, we randomly select 8 samples for each label.

For English datasets, we conduct \textsc{PT} based on T5-XXL with 11B parameters because previous works~\cite{prompt_tuning, cpm-v2} have shown that, T5-XXL is comparable with \textsc{FT} under the full-data setting. We also evaluate \textsc{FT} on various sizes of T5 to verify that larger models perform better and thus improving \textsc{PT} based on T5-XXL is meaningful. For Chinese datasets, we do \textsc{PT} based on a 11B model CPM-2. Since CPM-2 does not provide other size models, we compare it with mT5~\cite{mt5} of various sizes. 

Consistently, we use 100 soft tokens for \textsc{PT}. As a result, the tunable parameters is only $100 \times 4096 = 4.1\times10^5 = 410\text{K}$. Compared with the 11B ($1.1\times 10^{10}$) parameters of \textsc{FT}, \textsc{PT} only needs to store 3000 times smaller parameters for each task.

For prompt pre-training, we sample 10GB data from OpenWebText~\cite{openwebtext} for English tasks and 10GB data from WuDaoCorpora~\cite{wudaocorpora} for Chinese tasks. We use the Yelp-5~\cite{yelp} dataset to train the $\text{RoBERTa}_{\text{BASE}}$ model mentioned in Section \ref{sec:ssc}. More details of the training hyper-parameters can be found in the Appendix \ref{app:training_detail}.

\subsection{Main Results}
The main results of English and Chinese datasets are shown in Table \ref{tab:exp_main}. In the block \textsc{FT}, we present the \textsc{FT} results of the T5 model from the size small to XXL. In the block \textsc{PT}, we show the results of \textsc{PPT} and other baselines. The first baseline is Vanilla \textsc{PT}, where the soft prompts are randomly initialized from a normal distribution. The second is the hybrid strategy in Section \ref{sec:hybrid_pt}. We also consider LM Adaption used in \citet{prompt_tuning} in which the T5 model is further pre-trained for 10K steps with language modeling to reduce the gap between the pre-training and \textsc{PT}. We test two variants of \textsc{PPT}: Hybrid \textsc{PPT}, in which carefully designed hard prompts are combined with pre-trained soft prompt, and Unified \textsc{PPT}, in which all tasks are unified in the multiple-choice classification format.

\paragraph{Effectiveness}
From the Table \ref{tab:exp_main}  we have four observations. First, larger models achieve better overall performance, which means increasing the model size still helps under the few-shot setting. Therefore, we study \textsc{PT} on the large-scale pre-trained model. Note that for Chinese experiments, CPM-2 and mT5-XXL share the same parameter scale. Since CPM-2 outperforms mT5-XXL across all tasks, we use CPM-2 as the base model. 

Second, \textsc{PPT} outperforms Vanilla \textsc{PT} and LM Adaption on most datasets significantly. Although PPT is worse than Hybrid \textsc{PT} on BoolQ, combining \textsc{PPT} and hard prompts (Hybrid PPT) outperforms all baselines. This means pre-training soft prompts and using hybrid prompts are complementary. Similar phenomenons are observed on other datasets like RACE-m, LCQMC, and C$^3$, where adding hard prompts to \textsc{PPT} continues to improve results.

Third, \textsc{PPT} outperforms \textsc{FT} on all Chinese datasets and most English datasets. This indicates that there still remains a gap between masked language modeling and downstream tasks. Prompt pre-training bridges this gap to some extend. Based on this observation, an intuitive extension of our method is to further pre-train the entire model with $\text{PVP}^{\text{pre}}_i$ and fine-tune the model to the corresponding downstream tasks. However, since we focus on \textsc{PT} in this paper, we leave this as future work.

Fourth, \textsc{PPT} results in lower variances on most of the datasets. Few-shot learning is notorious for its instability, which becomes very obvious in Vanilla \textsc{PT}. For some datasets like SST-2, the variance reaches 15.5 which means the model does not perform better than random guesses under some random seeds. Combining with hard prompt or further pre-training with language modeling can alleviate this problem to some extent. But on some datasets like CCPM, Hybrid PT increases the variance and LM Adaption does not guarantee the average performance. With the help of pre-training, the variance remains at a low level across all datasets.

\paragraph{Unified PPT}

Unifying all formats to multiple-choice classification format is another variant of \textsc{PPT}. In Table \ref{tab:exp_main}, we can see that Unified \textsc{PPT} reaches comparable performance as \textsc{PPT} and Hybrid \textsc{PPT}, still outperforming other \textsc{PT} baselines. However, the datasets we have considered so far have no more than 5 labels. For tasks with more labels, especially single-text classification where pseudo label pre-training is not appropriate for cross-domain adaption, Unified \textsc{PPT} is a good alternative. In Table \ref{tab:exp-uni}, we test Unified \textsc{PPT} on datasets with more than 5 labels. 
\begin{table}[t]
    \centering
    \small
    \begin{tabular}{@{}lcc@{}}
        \toprule
        & TNews        & YahooAns  \\ \midrule
        $n_{\text{class}}$ & 14 & 10 \\
        FT                   & $43.2_{0.6}$ & $64.1_{1.9}$ \\
        PT                   & $41.2_{6.2}$ & $62.0_{4.2}$ \\
        PT (MC)              & $11.8_{2.1}$ & $60.8_{3.9}$ \\
        Unified PPT          & \bm{$50.6_{0.7}$} & \bm{$70.5_{1.9}$} \\ \bottomrule
    \end{tabular}
    \caption{The experiments on single-text classification tasks with more than 5 labels. Different from previous experiments, we randomly select 8 samples for each label. PT (MC) means doing \textsc{PT} in a multiple-choice format without prompt pre-training.}
    \label{tab:exp-uni}
\vspace{-2em}
\end{table}
For \textsc{PT} and \textsc{FT}, we use a verbalizer to map the labels to the intuitively selected words. \textsc{PT} (MC) means we solve the task in a multiple-choice classification format without prompt pre-training. We do not use \textsc{PPT} for single-sentence classification discussed in Section \ref{sec:ssc} because it is hard to find other suitable datasets to train the pseudo label annotator. However, we can see that Unified \textsc{PPT} still achieves the best performance, even exceeding \textsc{FT} by a large margin.

\begin{figure}[t]
    \includegraphics[width=\linewidth]{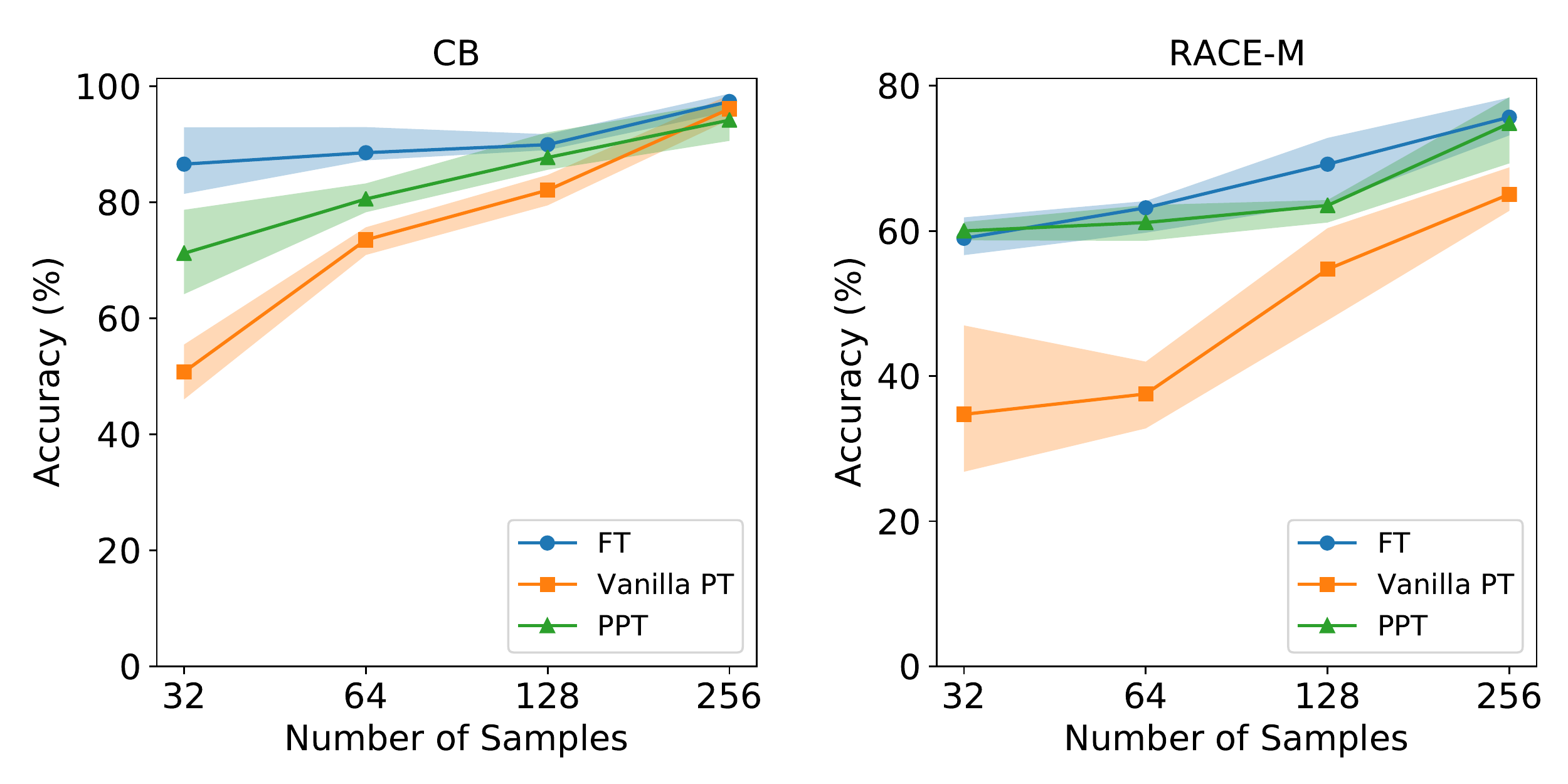}
    \caption{Comparison between FT, Vanilla PT, and PPT when different numbers of training samples are available. For the small number of samples, PPT is consistently better than Vanilla PT. When the number grows, the performance of these methods becomes closer.}\label{fig:exp_num}
\end{figure}

\subsection{Sample Efficiency}

\begin{table}[t]
\centering
\small
\begin{tabular}{@{}lc|ccc@{}}
\toprule
       & FT & PT & PPT &  Unified PPT\\ \midrule
SST-2  & $96.1_{0.2}$  & $96.8_{0.1}$ & $96.9_{0.1}$ & \bm{$97.0_{0.1}$} \\
SST-5  & $58.4_{1.4}$  & $58.5_{1.1}$ & \bm{$59.3_{1.2}$} & $58.3_{0.2}$ \\
RACE-m & $86.8_{1.4}$ & $85.0_{0.5}$ & $85.9_{0.4}$ & \bm{$86.4_{0.6}$} \\
RACE-h & $83.7_{0.6}$ & $82.5_{1.9}$ & $83.9_{1.3}$ & \bm{$84.3_{0.5}$} \\
BoolQ  & $90.9_{0.6}$ & $89.4_{0.6}$ & $89.3_{0.3}$ & \bm{$89.4_{0.3}$} \\
RTE     & $89.8_{1.0} $ & $88.0_{4.8}$ & $89.6_{0.8}$ & \bm{$91.8_{0.7}$} \\
CB      & $94.6_{1.2}$ & \bm{$94.3_{5.6}$} & $93.7_{3.1}$ & $92.9_{4.9}$ \\ \bottomrule
\end{tabular}
\caption{The performance of FT, PT, PPT, and Unified PPT when the full training datasets are available. We report the mean and the standard deviation over 3 random seeds on the validation set.}\label{tab:full_data}
\end{table}

We discuss how the performance of FT, PT, and PPT varies when the number of training samples increases. In Figure \ref{fig:exp_num}, we show the trend of these methods on the RACE-m and CB datasets. For 32 to 128 samples, PPT is consistently better than PT, and the performances of the three methods gradually converge when the number grows to 256.

\begin{figure}[t]
    \includegraphics[width=\linewidth]{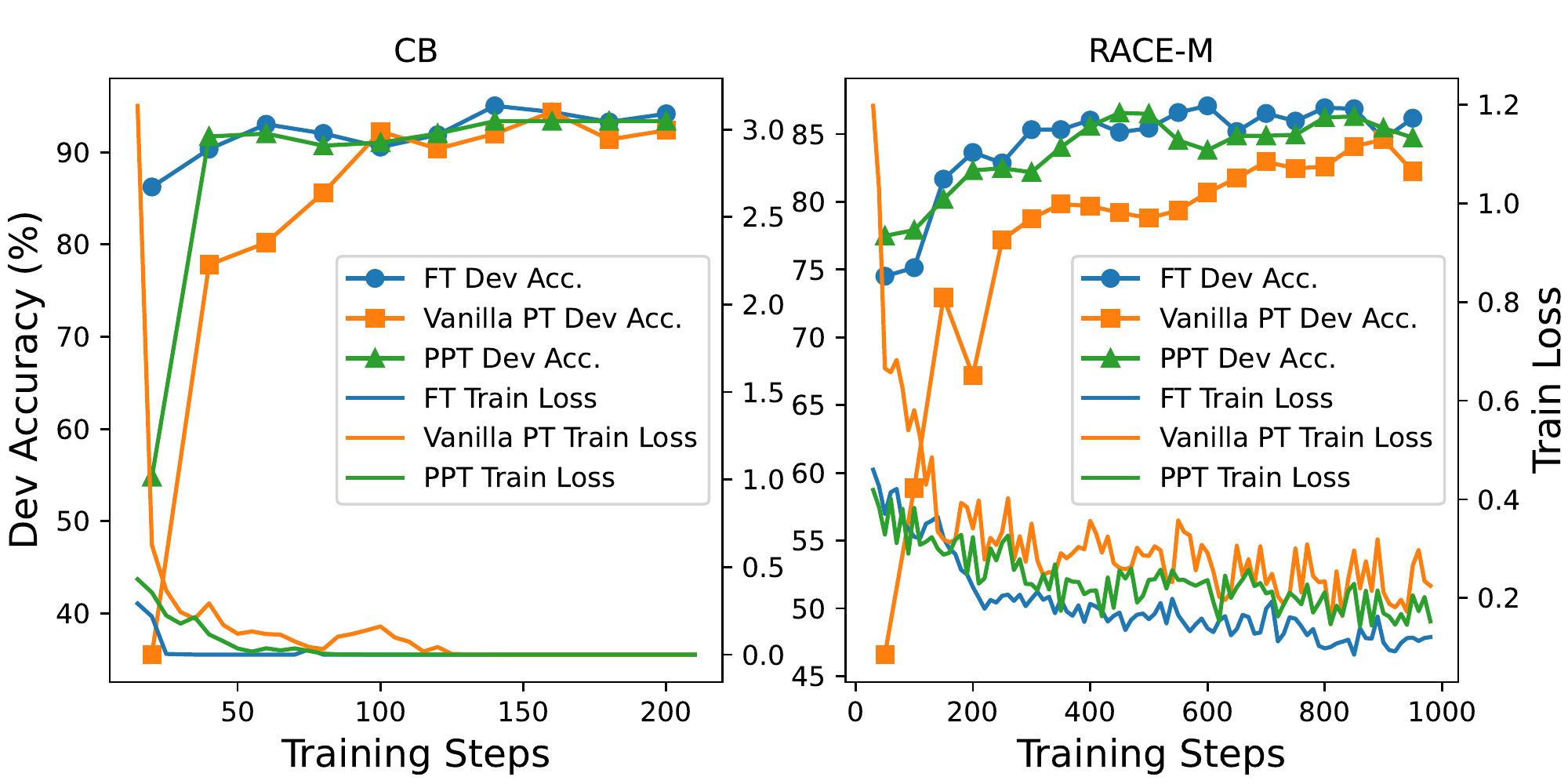}
    \caption{Comparison of the convergence between FT, Vanilla PT, and PPT. PT converges much slower than FT. Owing to the pre-trained initialization, PPT significantly speeds up the convergence.}\label{fig:conv}
\end{figure}

We also compare different tuning approaches given the full training data. From Table \ref{tab:full_data}, we can see that PPT and Unified PPT still outperform the Vanilla PT on most datasets. In addition, we observe that although PT is faster than FT in a single optimization step, it converges much slower, which results in an even longer training time. We argue that PPT can be an effective solution to this problem. As shown in Figure \ref{fig:conv}, with the pre-trained initialization, PPT speeds up the convergence of Vanilla PT on both RACE-m and CB datasets. We give a more detailed analysis of the training consumption in the Appendix \ref{sec:consumption}. Since PPT still converges a bit slower than FT, how to further accelerate the convergence of PT is worth studying in future work.

\section{Related Works}

\paragraph{PLMs and Task-oriented Fine-tuning} Recently, various powerful PLMs have been proposed, such as GPT~\cite{gpt}, BERT~\cite{bert}, RoBERTa~\cite{roberta} and T5~\cite{t5}. To adapt these PLMs to downstream NLP tasks, task-oriented fine-tuning has been proposed, where researchers use PLMs as the backbone and add some task-specific heads to optimize task-specific objectives. Then, all parameters of both PLMs and additional heads are tuned using task-specific data. Results have shown that task-oriented fine-tuning can outperform models trained from scratch on a series of NLP tasks. 

\paragraph{Prompt-oriented Fine-tuning} Most existing PLMs are pre-trained with language modeling objectives, yet the objectives of downstream tasks are quite different. To overcome the gap between pre-training and downstream tasks, prompt-oriented fine-tuning is introduced. In prompt-oriented fine-tuning, downstream tasks are also formalized as language modeling problems by inserting language prompts, and the results of language modeling can correspond to the solutions of downstream tasks. 

Knowledge probing~\cite{lama,trinh2018simple,davison2019commonsense} is the seminal work that stimulates the development of prompts. In knowledge probing, language triggers are widely used to induce PLMs to generate relational facts. These pioneering works demonstrate that language prompts can effectively stimulate the knowledge from PLMs. Encouraged by this, manually designing hard prompts consisting of discrete words is first used in prompt-oriented fine-tuning~\citet{pet, pet2}. Considering manually designing prompts is both time-consuming and difficult to find the best choice, later works~\cite{lm-bff,how-can-we-know, autoprompt} proposed to generate prompts automatically. However, these works still restrict auto-generated prompts to discrete spaces which are usually sub-optimal. 

To overcome the shortcomings of discrete spaces, \citet{prefixtuning,p-tuning,ptr,warp,optiprompt} explore to combine hard prompts and soft prompts. Different from hard prompts using concrete and discrete tokens, soft prompts are composed of several continuous learnable embeddings, and these embeddings are randomly initialized. To step forward, some works~\cite{prefixtuning,learning-how-to-ask,prompt_tuning} propose to only tune soft prompts and fix the entire PLM parameters. When models are large enough, this method can be comparable to full-model tuning.

\paragraph{Few-shot Learning with PLMs} 
Since long-tail distribution is common in real-world applications, few-shot learning is quite meaningful for the stable and effective use of PLMs, thereby attracts much attention recently. Apart from GPT-3~\cite{gpt3} and PET\cite{pet} which demonstrates the superiority of PLMs in few-shot scenarios, some later works~\citet{true-few-shot, flex} also discuss reasonable few-shot settings by restricting the size of validation set and proposing a unified framework to evaluate few-shot performance. There is also work~\cite{cutting-down-prompt} pointing out the low performance of \textsc{PT} for few-shot learning. But they mostly focus on PLMs with fewer than 400M parameters. In this paper, we study few-shot learning on large-scale 11B PLMs.


\section{Conclusion and Future Work}

In this paper, we present \textsc{PPT}, a framework that improves prompt tuning for few-shot learning. We propose to firstly unify downstream tasks to several formats. Then, we design self-supervised pre-training tasks for each format and pre-train prompts on these tasks. Finally, we do prompt tuning on downstream tasks based on the pre-trained initialization. Extensive experiments show that our method significantly outperforms other prompt tuning baselines, performing comparable or even better than full-model tuning. 

There are three important directions for future work: (1) Designing unified task formats and the corresponding pre-training objectives for other kinds of tasks such as language generation and relation extraction. (2) Evaluating the few-shot performance of other parameter-efficient tuning approaches~\cite{unideta} and adapting unified task pre-training to them. (3) Beyond the soft prompt, studying whether unified task pre-training helps the pre-trained language models itself.

\section*{Acknowledgements}

This work was supported by the National Science Foundation for Distinguished Young Scholars (with No. 62125604) and the NSFC projects (Key project with No. 61936010 and regular project with No. 61876096). This work was also supported by the Guoqiang Institute of Tsinghua University, with Grant No. 2019GQG1 and 2020GQG0005.

\bibliography{anthology}
\bibliographystyle{acl_natbib}

\appendix
\section*{Appendices}
\section{Dataset Information}
\label{app:dataset}

Since some of the test sets of the datasets we used is not publicly available, we follow \citet{few-sample-bert} and \citet{lm-bff} to use original validation sets for testing. For English experiments, we use a dataset from GLUE~\cite{glue} (SST-2~\cite{sst-2}), datasets from SuperGLUE~\cite{superglue}, (BoolQ~\cite{boolq}, CB~\cite{cb}, and RTE~\cite{rte}), two extra single-text classification datasets (SST-5~\cite{sst-2} and YahooAnswers~\cite{yahoo-ans}), and two standard question answering datasets (RACE-middle and RACE-high)~\cite{race} for multiple-choice classification. For Chinese experiments, we use four datasets from CLUE~\cite{clue} (CMNLI\footnote{\url{https://www.cluebenchmarks.com/}\label{foot:clue}} , OCNLI~\cite{ocnli}, TNews\footref{foot:clue}, C$^3$~\cite{c3}), two sentiment analysis datasets (ChnSent\footnote{\url{https://github.com/SophonPlus/ChineseNlpCorpus}\label{foot:chn}} and Amazon Reviews\footref{foot:chn}), and one extra natural language inference dataset LCQMC~\cite{lcqmc}.

\section{PVPs for Chinese Tasks}

We describe the $\text{PVP}^\text{pre}_i$ for Chinese datasets in this section. Just like English scenarios, all these \text{PVP}s are simple and intuitive. 

\paragraph{Sentence-Pair Classification}
Given the input $\bm{x} = (\bm{s}_1, \bm{s}_2)$, the label list $\mathcal{Y} = [0, 1, 2]$, we have:
\begin{equation}
    \small
    \begin{aligned}
        f^{\text{pre}}_i(\bm{x}) &= \text{``}\bm{s}_1 \left\langle \text{X} \right\rangle \text{。} \bm{s}_2\text{''}, \\
        v^{\text{pre}}_i(\mathcal{Y}) &= [\text{矛盾}, \text{中立}, \text{相似}].            
    \end{aligned}
\end{equation}

\paragraph{Multiple-Choice Classification}

Given a input $\bm{x}$ consisting of a query and six candidates: $\bm{x} = (\bm{s}_q, \bm{s}_1, \bm{s}_2, \cdots , \bm{s}_6)$, we convert $x$ to a language sequence by defining the $\textsc{PVP}^\text{pre}_i$ as follows:
\begin{equation}
    \small
    \begin{aligned}
        f^{\text{pre}}_i(\bm{x}) &= \text{``}\bm{s}_q\text{？一、}\bm{s}_1 \cdots \text{六、}\bm{s}_6. \text{答案是} \left\langle \text{X}\right\rangle\text{。''}, \\
        v^{\text{pre}}_i(\mathcal{Y}) &= [\text{一}, \text{二}, \text{三}, \text{四}, \text{五}, \text{六}].
    \end{aligned}    
\end{equation}

\paragraph{Single-Sentence Classification}
Similar to the English scenario, we take sentiment classification as an example. Given the input $\bm{x} = (\bm{s})$, we have:
\begin{equation}
    \small
    \begin{aligned}
        f^{\text{pre}}_i(\bm{x}) &= \text{``}\bm{s}\text{。}\left\langle \text{X}\right\rangle\text{。''}, \\
        v^{\text{pre}}_i(\mathcal{Y}) &= [\text{差}, \text{不好}, \text{一般}, \text{好}, \text{赞}].
    \end{aligned}    
\end{equation}

Based on the $\textsc{PVP}^\text{pre}_i$, the design of $\textsc{PVP}^k_i$ is similar to that of English tasks.

\begin{table}[t]
    \centering
    \small
    \begin{tabular}{@{}ll@{}}
        \toprule
        \multicolumn{2}{c}{English} \\ \midrule
        SPC & $\bm{P}$ Question: $\bm{s}_1$ ? $\left\langle \text{X}\right\rangle$. $\bm{s}_2$  \\
        MCC & $\bm{P}$ We ask $\bm{s}_q$ ? $\text{A.}\bm{s}_1 \cdots \text{F.}\bm{s}_6. $The answer is $\left\langle \text{X}\right\rangle$.        \\
        SSC & $\bm{P}$ $\bm{s}$. It was $\left\langle \text{X}\right\rangle$. \\ \midrule
        \multicolumn{2}{c}{Chinese} \\ \midrule
        SPC & $\bm{P}$ 问题：$\bm{s}_1$？$\left\langle \text{X}\right\rangle$。 $\bm{s}_2$        \\
        MCC & $\bm{P}$ 问题：$\bm{s}_q$？$\text{一、}\bm{s}_1 \cdots \text{六、}\bm{s}_6$.答案是：$\left\langle \text{X}\right\rangle$。 \\
        SSC & $\bm{P}$ $\bm{s}$。这很 $\left\langle \text{X}\right\rangle$。 \\ \bottomrule
    \end{tabular}
    \caption{The hard prompts for Hybrid \textsc{PT} and Hybrid \textsc{PPT}. SSC stands for single-sentence classification, MCC stands for multiple-choice classification, and SPC stands for sentence-pair classification.}
    \label{tab:hard}
\end{table}

\section{Training Details}
\label{app:training_detail}

Considering the instability of the few-shot learning, we run each experiment 5 times on the random seed [10, 20, 30, 40, 50] and report the averaged performance as well as the standard deviation. Due to the resource limit, for 11B models, we adopt model parallelism~\cite{megatron} and store a model with 4 GPU devices. We also use mixed-precision training~\cite{mixed-fp} and ZeRO~\cite{zero} stage-1 provided in DeepSpeed~\cite{deepspeed} to reduce GPU memory usage. For models in other sizes, we all use full-precision training. We describe the details of the training hyper-parameters in the following sections. 

\subsection{Full-Model Tuning}

For Full-Model Tuning (\textsc{FT}), we tune the entire parameters of the model without concatenating soft prompts. For all models, we fix the batch size as 16. In this way, we train the largest 11B model with 16 NVIDIA V100 32G GPUs. We find that different sized models prefer significantly different learning rates. Therefore, we search for the learning rates in varied intervals and show each model size and its corresponding searching interval in Table \ref{tab:lr}. We train the model for 50 epochs and do evaluation every 6 optimization steps. We choose the model performing the best on the validation set and evaluate it on the test set. 

\begin{table}[t]
    \centering
    \small
    \begin{tabular}{c|c}
    \toprule
        Model Size & Searching Interval \\
    \midrule
        Small & 2e-4, 5e-4, 1e-3\\
        Base  & 2e-4, 5e-4, 1e-3\\
        Large & 5e-5, 1e-4, 2e-4\\
        XL    & 3e-5, 5e-5, 1e-4\\
        XXL   & 3e-6, 5e-6, 1e-5\\
    \bottomrule
    \end{tabular}
    \caption{The searching intervals of learning rates for the models with different sizes. Generally, small models prefer large learning rates. }
    \label{tab:lr}
\end{table}

\subsection{Prompt Tuning}

For Prompt Tuning (PT), we add a set of soft prompts before the input text. When adapting the model to downstream tasks, we only tune the soft prompts with the entire model fixed. Similar to \textsc{FT}, we fix the batch size as 16 and train the model for 50 epochs, while evaluating the model every 6 steps. Since the tunable parameters are much less in \textsc{PT}, 8 NVIDIA V100 32G GPUs are enough for the training. We find \textsc{PT} requires a much larger learning rate than \textsc{FT}. Therefore, we search for the learning rate in [5e-3, 1e-2, 2e-2, 5e-2] and choose the model with the best performance on the validation set. This observation also implies that \textsc{PT} is much harder to train than \textsc{FT}, which is consistent with the experiment results in the main paper.

\subsection{Prompt Pre-Training}
We use the sampled 10GB data to construct the pre-training data for each task format for prompt pre-training. Across all tasks, we use the ``inverse square root'' learning rate scheduler~\cite{t5} and set the learning rate in this scheduler as 0.1 with no warmup steps. We set the batch size as 256, the max input length as 512, and train the prompts for at most 200,000 steps. We split 5\% data for validation and the rest for pre-training. We evaluate the performance on the validation set every 2,000 steps and choose the prompt with the lowest validation loss. The details of constructing the pre-training data for each task are as follows.

\paragraph{Sentence-Pair Classification} In the next sentence prediction task, we set the two sentences next to each other as label 2, those from the same document but not true next sentence as 1, and those from different documents as 0. We filter out the sentences with less than 5 tokens and the pairs in which the two sentences' length ratios are larger than 100. 

\paragraph{Multiple-Choice Classification} In the next sentence selection task, giving a query sentence, the options contain one adjacent sentence, one sentence from the same document as the query, and four from the different documents. We also filter out the sentences with less than 5 tokens. To fit in the max input length, we truncate the query sentence to 389 tokens and the options to 86 tokens. For Unified PPT, we uniformly sample the option numbers from 2 to 16 to cover more downstream circumstances. The input configurations of different option numbers is shown in Table \ref{tab:unify_detail}. 

\begin{table}[t]
    \centering
    \small
    \begin{tabular}{cccccc}
    \toprule
       Num.  & len(q) & len(op) & Pos. & Neg.-S & Neg.-D \\ \midrule
       2     & 400       &  50       &  1   &  1     &  0   \\
       3     & 400       &  50       &  1   &  1     &  1   \\
       4     & 400       &  50       &  1   &  1     &  2   \\
       5     & 400       &  40       &  1   &  1     &  3   \\
       6     & 300       &  40       &  1   &  1     &  4   \\
       7     & 250       &  30       &  1   &  2     &  4   \\
       8     & 200       &  30       &  1   &  2     &  5   \\
       9     & 200       &  30       &  1   &  2     &  6   \\
      10     & 150       &  20       &  1   &  2     &  7   \\
      11     & 150       &  20       &  1   &  3     &  8   \\
      12     & 150       &  20       &  1   &  3     &  9   \\
      13     & 150       &  20       &  1   &  3     &  10  \\
      14     & 150       &  20       &  1   &  3     &  11   \\
      15     & 150       &  20       &  1   &  3     &  12   \\
      16     & 150       &  20       &  1   &  3     &  13   \\ \bottomrule
    \end{tabular}
    \caption{The input configurations of different option numbers. ``Num.'' means the number of the options. ``len(q)'' and ``len(op)'' means the maximum length of the query and the options. ``Pos.'' means the number of positive options. ``Neg.-S'' and ``Neg.-D'' represent the negative options from the same and different documents.}
    \label{tab:unify_detail}
\end{table}

\paragraph{Single-Sentence Classification} We use the $\text{RoBERTa}_\text{BASE}$ model trained on the Yelp-5 dataset to annotate pseudo labels on the unlabeled data. We use learning rate 1e-4, batch size 16, warm-up rate 0.01, and train the model for 10 epochs. We choose the checkpoint with the highest accuracy on the validation set, which is 70.53 at the 5-th epoch, to annotate the label. We set different minimal classification confidence thresholds for the 5 labels to control annotation quality and balance the label. The thresholds of the label 0 $\sim$ 4 are [0.95, 0.50, 0.50, 0.50, 0.70].

\section{Hard Prompts}

In this section, we describe the hard prompts we use in Hybrid \textsc{PT} and Hybrid \textsc{PPT}. For simplicity, we choose the best hard prompts for each task format (e.g. sentence-pair classification, multiple-choice classification, and single-sentence classification) based on \textsc{PT}  in pilot experiments and directly use them in Hybrid \textsc{PPT}. The hard prompts corresponding to each task format are shown in Table \ref{tab:hard}.

\begin{table*}[t]
\small
\centering
\begin{tabular}{@{}llrrrrrrr@{}}
\toprule
                    &                       & SST-2 & SST-5 & RACE-m & RACE-h & BoolQ & RTE  & CB    \\ \midrule
\multirow{2}{*}{FT} & Single Step Time (ms) & 4,416   &  4,419  & 6,498   &  6,238   &  4,760   & 4,653  & 5,962   \\
                    & GPU Mem. Cost (GB)    &  259  & 259  & 512  &   512  &  314   &  346    & 512    \\ \midrule
\multirow{2}{*}{PT} & Single Step Time (ms) & 794   & 791   & 4,000  & 3,976  & 1,089 & 944  & 1,655  \\
                    & GPU Mem. Cost (GB)         & 72  & 72  & 159  & 154  & 82  & 81 & 102 \\ \bottomrule
\end{tabular}
\caption{The time cost for a single optimization step and GPU memory usage throughout the training. PT has a shorter single-step optimization time and a lower GPU memory cost.}\label{tab:consumption}
\end{table*}

\section{Training Consumption}
\label{sec:consumption}
We analyze the time and memory consumption of FT and PT in this section. For PPT, the consumption is exactly the same as PT during the downstream adaption. Although pre-training prompts may introduce external costs, we only need to do it once and use the pre-trained prompts for multiple tasks. From Table \ref{tab:consumption}, we can see that PT's optimization time of a single step is much shorter than FT, and it occupies much less GPU memory. The reason is that during optimization, PT only needs to update the prompt parameters, which means the momentum and gradients of other parameters are not required to be stored and transmitted to between different GPU devices.

\label{app:chinese}



\end{CJK*}
\end{document}